\documentclass[sigconf,natbib,screen=true,anonymous=false,review=false]{acmart}

\usepackage{booktabs} 
\usepackage{algorithm, algpseudocode}
\usepackage{graphics}
\usepackage{epsfig}
\usepackage{graphicx}
\usepackage{xcolor}
\usepackage{subfigure}
\usepackage{balance}
\usepackage{mathrsfs}
\usepackage{multirow}
\usepackage{tabularx}
\usepackage{rotating}
\usepackage{float}
\usepackage{pifont}
\usepackage{CJKutf8}
\usepackage[inline]{enumitem}
\usepackage{ulem}
\usepackage{acronym}
\usepackage[skip=0pt]{caption}

\newcommand{\cmark}{{\color{green} \ding{52}}}

\newcommand{\xmark}{{\color{red} \ding{56}}}

\AtBeginDocument{%
  \providecommand\BibTeX{{%
    \normalfont B\kern-0.5em{\scshape i\kern-0.25em b}\kern-0.8em\TeX}}}

\DeclareMathOperator*{\softmax}{softmax}

\newcommand{\OurMethod}{VRBot}

\acrodef{CDS}{clinical decision support}
\acrodef{ELBO}{evidence lower bound}
\acrodef{KGC}{knowledge grounded conversation}
\acrodef{MDG}{medical dialogue generation}
\acrodef{MDS}{medical dialogue system}
\acrodef{MRG}{medical response generation}
\acrodef{PPL}{physician policy learning}
\acrodef{PST}{patient state tracking}
\acrodef{SGVB}{stochastic gradient variational Bayes}
\acrodef{TDS}{task-oriented dialogue system}

\newcommand{\OurParagraph}[1]{\smallskip\noindent\textbf{#1}}

\newcommand{\smallnegskip}{\vspace*{-1mm}}
\newcommand{\negskip}{\vspace*{-2mm}}

\copyrightyear{2021}
\acmYear{2021}
\setcopyright{acmlicensed}\acmConference[SIGIR '21]{Proceedings of the 44th International ACM SIGIR Conference on Research and Development in Information Retrieval}{July 11--15, 2021}{Virtual Event, Canada} \acmBooktitle{Proceedings of the 44th International ACM SIGIR Conference on Research and Development in Information Retrieval (SIGIR '21), July 11--15, 2021, Virtual Event, Canada}
\acmPrice{15.00}
\acmDOI{10.1145/3404835.3462921} 
\acmISBN{978-1-4503-8037-9/21/07}

\ccsdesc[100]{Applied computing~Health care information systems}
\ccsdesc[100]{Computing methodologies~Discourse, dialogue and pragmatics}
\ccsdesc[300]{Information systems~Specialized information retrieval}
\ccsdesc[300]{Information systems~Users and interactive retrieval}

\keywords{Medical dialogue systems; Task-oriented dialogue generation; Variational inference; Semi-supervised learning}

\author{Dongdong Li$^{1}$*\qquad Zhaochun Ren$^{1\dagger}$*\qquad Pengjie Ren$^1$\qquad Zhumin Chen$^1$}
\author{
Miao Fan$^2$\qquad Jun Ma$^1$ \qquad Maarten de Rijke$^{3,4}$
}

\makeatletter
\def\authornotetext#1{
\if@ACM@anonymous\else
    \g@addto@macro\@authornotes{
    \stepcounter{footnote}\footnotetext{#1}}
\fi}
\makeatother
\authornotetext{Equal contribution.}
\authornotetext{Corresponding author.}

\affiliation{
 \institution{\textsuperscript{\rm 1}Shandong University, Qingdao \country{China}}
  \institution{\textsuperscript{\rm 2}Baidu Inc., Beijing \country{China}}
 \institution{\textsuperscript{\rm 3}University of Amsterdam, Amsterdam \country{The Netherlands}}
 \institution{\textsuperscript{\rm 4}Ahold Delhaize, Zaandam \country{The Netherlands}}
}

\email{lddsdu@gmail.com, {zhaochun.ren, chenzhumin, majun}@sdu.edu.cn} 
\email{jay.ren@outlook.com, fanmiao@baidu.com, M.deRijke@uva.nl}

\def\authors{Dongdong Li, Zhaochun Ren, Pengjie Ren, Zhumin Chen, Miao Fan, Jun Ma, and Maarten de Rijke}

\begin{document}
\begin{sloppypar}
\fancyhead{}

\title[Semi-Supervised Variational Reasoning for Medical Dialogue Generation]{Semi-Supervised Variational Reasoning for\\ Medical Dialogue Generation}


\begin{abstract}
Medical dialogue generation aims to provide automatic and accurate responses to assist physicians to obtain diagnosis and treatment suggestions in an efficient manner.
In medical dialogues two key characteristics are relevant for response generation: \emph{patient states} (such as symptoms, medication) and \emph{physician actions} (such as diagnosis, treatments).
In medical scenarios large-scale human annotations are usually not available, due to the high costs and privacy requirements.
Hence, current approaches to medical dialogue generation typically do not explicitly account for patient states and physician actions, and focus on implicit representation instead.

We propose an end-to-end variational reasoning approach to medical dialogue generation. 
To be able to deal with a limited amount of labeled data, we introduce both patient state and physician action as latent variables with categorical priors for explicit \textit{patient state tracking} and \textit{physician policy learning}, respectively. 
We propose a variational Bayesian generative approach to approximate posterior distributions over patient states and physician actions. We use an efficient stochastic gradient variational Bayes estimator to optimize the derived evidence lower bound, where a 2-stage collapsed inference method is proposed to reduce the bias during model training. 
A physician policy network composed of an action-classifier and two reasoning detectors is proposed for augmented reasoning ability.
We conduct experiments on three datasets collected from medical platforms. 
Our experimental results show that the proposed method outperforms state-of-the-art baselines in terms of objective and subjective evaluation metrics. Our experiments also indicate that our proposed semi-supervised reasoning method achieves a comparable performance as state-of-the-art fully supervised learning baselines for physician policy learning.
\end{abstract}

\maketitle
\acresetall


\negskip
\section{Introduction}

\begin{figure}
  \centering
  \includegraphics[width=0.85\linewidth]{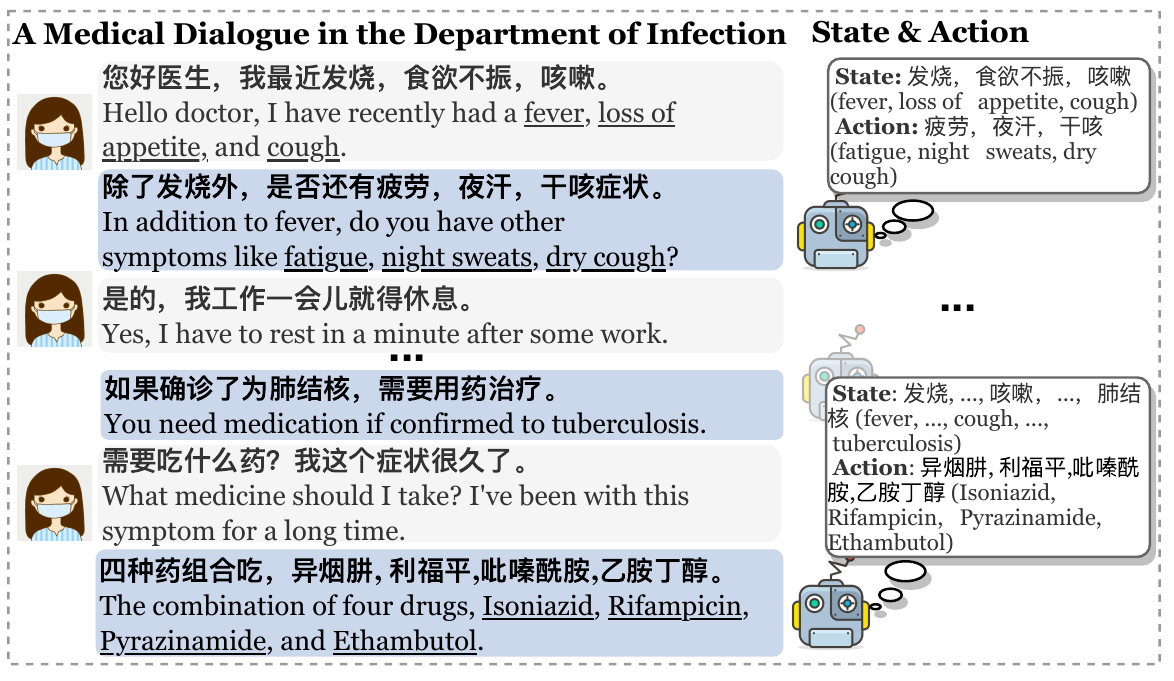}
  \caption{An example medical dialogue in the infection department, the left part shows the dialogue; the right part illustrates dialogue states and actions.}
  \label{fig:intro-demo}
\end{figure}
Increasingly, conversational paradigms are being used to connect people to information, both to address open domain information needs \citep[e.g.,][]{lei2018sequicity,jin2018explicit,qu-2020-open-domain,lei2020re,vakulenko-2020-analysis,hua-2020-learning,lei-2020-conversational} and in support of professionals in highly specialized vertical domains \citep[e.g.,][]{zhang-2020-summarizing,terhoeve-2020-conversations}. 
Our focus is on conversational information seeking approaches in the medical domain.
During clinical treatment, a conversational medical system can serve as a physician's assistant to help generate responses for a patient's need, i.e, inquire about symptoms, make a diagnosis, and prescribe medicine or treatment~\citep{wei2018task,xu2019end,xia2020generative}. 
Intelligent \acp{MDS} are able to reduce the work pressure of physicians~\cite{shi2020understanding}. 
Previous work on \acp{MDS} mostly focuses on producing an accurate diagnosis given the dialogue context~\cite{wei2018task,xia2020generative,xu2019end,lin2019enhancing}.
There is very little work that considers the task of multi-turn medical dialogue generation to provide proper medical responses by tapping into large-scale medical knowledge sources.

There are two key characteristics that are specific to \ac{CDS}, and hence for dialogue systems that are meant to support clinical decision making: \emph{patient states} (e.g., symptoms, medicine, etc.) and \emph{physician actions} (e.g., treatments, diagnosis, etc.).
These two characteristics make \acp{MDS} more complicated than other knowledge-intensive dialogue scenarios.
Similar to \acp{TDS}, a \ac{MDG} process can be decomposed into $3$ stages:  
\begin{enumerate*}[label=(\arabic*)]
\item \acfi{PST}: after encoding the patient's descriptions, the \ac{MDS} tracks the patient's physiological condition, i.e., \textit{patient states}, in the discourse context; 
\item \acfi{PPL}: given the patient's states and utterances, the \ac{MDS} generates the \textit{physician's action} to embed into the response;  and
\item \ac{MRG}: the \ac{MDS} responds with a coherent sentence based on detected states and actions. 
\end{enumerate*}

Figure~\ref{fig:intro-demo} shows an example medical dialogue from the infection department. The left part lists the conversation, whereas the right part indicates patient states and physician actions during the conversation.
At the first turn the patient shares their symptoms, i.e., \textit{fever}, \textit{loss of appetite}, and \textit{cough}, as the patient state; the physician asks if the patient has other symptoms, i.e., \textit{fatigue}, \textit{night sweats}, and \textit{dry cough}, to reflect the physician action at the second turn. Both states and actions vary as the conversation develops.
At the last turn, the physician's action is to prescribe drugs: \textit{Isoniazid}, \textit{Rifampicin}, \textit{Pyrazinamide}, and \textit{Ethambutol}.

The development of end-to-end \ac{MDG} solutions faces a number of challenges:
\begin{enumerate*}[label=(\arabic*)]
\item Most \acp{TDS} need a large amount of manually labeled data to predict explicit dialogue states. In medical dialogues, annotators need medical expertise to annotate data. For privacy reasons, large-scale manually labeling intermediate states is problematic. Hence, few \ac{TDS} methods can directly be applied to \ac{MRG}~\citep{wen2016network}.
\item Existing approaches to \ac{MDG} have a limited semantic understanding of the domain, which makes it hard to generate knowledgeable responses in a medical context~\cite{liu2020meddg}. 
\item To help patients or physicians understand why a \ac{MDG} system generates a response, explainability with indicative and interpretable information is indispensable, which is ignored by most \ac{TDS} studies.
\end{enumerate*}

To address these challenges, we propose \OurMethod, which performs variational reasoning for \ac{MRG}.
Inspired by approaches to \ac{TDS}, \OurMethod{} contains a \emph{patient state tracker}  and a \emph{physician policy network} to detect patient states and physician actions, respectively. 
Unlike previous work, which learns from massive amounts of human-labeled observed variables, \OurMethod{} considers the patient state and the physician action as dual latent variables inferred in a variational Bayesian manner. 
We employ a \acf{SGVB} estimator to efficiently approximate the posterior inference. To alleviate the bias problem during \ac{SGVB} estimation, we propose a  2-stage collapsed inference method to iteratively approximate the posterior distribution over states and actions.

To address the problem of limited semantic understanding during response generation, we proceed as follows. The physician policy network comprises an \textit{action-classifier} that classifies physician actions into \emph{action categories}, and two reasoning components, a \textit{context reasoning detector} and a \textit{graph reasoning detector}, that infer explicit action keywords through the dialogue context and medical knowledge graph, respectively.
With explicit sequences of patient states, physician actions, and multi-hop reasoning, \OurMethod{} is able to provide a high degree of explainability of its medical dialogue generation results.

To assess the effectiveness of \OurMethod{}, we collect a knowledge-aware medical dialogue dataset, KaMed. KaMed contains over 60,000 medical dialogue sessions with 5,682 entities (such as \textit{Asthma} and \textit{Atropine}).
Using KaMed and two other \ac{MDG} benchmark datasets, we find that \OurMethod{}, using limited amounts of labeled data, outperforms state-of-the-art baselines for \ac{MDG}.
Hence, given large-scale unlabeled medical corpora, \OurMethod{} can accurately trace the patient's physiological conditions and provide more informative and engaging responses by predicting appropriate treatments and diagnosis.
We also find that \OurMethod{} is able to provide more explainable response generation process over other MDG baselines.

Our contributions are as follows:
\begin{enumerate*}[label=(\arabic*)]
    \item We propose an end-to-end medical response generation model, named \OurMethod{}. To the best of our knowledge, \OurMethod{} is the first to simultaneously model states and actions as latent variables in \acp{TDS}.
    \item We devise a hybrid policy network that contains a context-reasoning detector and a graph-reasoning detector, which allow \OurMethod{} to predict physician actions based on the dialogue session and external knowledge simultaneously.
    \item We show that \OurMethod{} can explicitly track patient states and physician actions even with few or no human-annotated labels.
    \item We release KaMed, a large-scale medical dialogue dataset with external knowledge.
    \item Experiments on benchmark datasets show that \OurMethod{} is able to generate more informative, accurate, and explainable responses than state-of-the-art baselines.
\end{enumerate*}


\negskip
\section{Related work}
\label{sec:related work}

\textbf{Medical dialogue systems.}
Previous methods for \acp{MDS} are modeled after \acp{TDS}, following the paradigm that a patient expresses their symptoms.
\citet{wei2018task} propose to learn a dialogue policy for automated diagnosis based on reinforcement learning. 
\citet{lin2019enhancing} build a symptom graph to model associations between symptoms to boost the performance of symptom diagnosis. 
\citet{xu2019end} consider the co-occurrence probability of symptoms with diseases explicitly with reinforcement learning. 
\citet{xia2020generative} improve upon this work using mutual information rewards and generative adversarial networks. 
Meanwhile, various approaches have been explored to improve the understanding of medical dialogue histories, including symptom extraction~\cite{du2019extracting}, medical slot-filling~\cite{shi2020understanding}, and medical information extraction~\cite{zhang2020mie}.
\citet{chen2020meddiag} investigate the performance of pre-trained models for predicting response entities.
\citet{chen2020meddiag} collect a dataset that consists of millions of dialogue sessions but do not explicitly consider learning the dialogue management as there are no human-annotated labels. 

Currently, no prior work is able to explicitly learn a dialogue policy from a large-scale unlabeled corpus, greatly limiting the application of medical dialogue systems. 

\OurParagraph{Dialogue state tracking.}
Dialogue state tracking plays an important role for \acp{TDS}. 
Conditional random field-based approaches~\cite{lee2013structured,lee2013recipe} and deep neural network-based approaches~\cite{henderson2013deep,mrkvsic2016neural} have been proposed to track states in modular \acp{TDS}~\cite{chen2017survey}.
Recently, end-to-end \acp{TDS} have attracted a lot of interest \cite{mehri2019structured,lei2018sequicity,wu2019alternating,hosseini2020simple,jin2018explicit,liao2020rethinking,zhang2020probabilistic}. 
For non-task-oriented dialogue generation, \citet{serban2016building} and \citet{chen2018hierarchical} propose generation methods with implicit state representations, for which it is hard to distinguish medical concepts. 
Dialogue states have also been represented as a sequence of keywords from the dialogue context~\cite{wang2018chat}.
\citet{jin2018explicit} and \citet{zhang2020probabilistic} propose semi-supervised generative models to leverage unlabeled data to improve state tracking performance. 
\citet{liang2020moss} propose an encoder-decoder training framework, MOSS, to incorporate supervision from various intermediate dialogue system modules. MOSS exploits incomplete supervision during model training.
However, existing approaches fail to generate engaging and informative responses as do not address the semantic reasoning ability of the dialogue agents.
As far as we know, no existing method simultaneously models states and actions under a few-shot regime.

In the \ac{MDG} scenario, learning physician actions is as important as state tracking. Compared to \cite{jin2018explicit, zhang2020probabilistic,liang2020moss}, our model is capable of inferring missing states and actions simultaneously.

\OurParagraph{Knowledge-grounded conversations.}
The task of \ac{KGC} is to generate responses based on accurate background knowledge. 
The task can be grounded into two categories according to the format of the background knowledge, i.e., structured \ac{KGC} and unstructured \ac{KGC}.
The former focuses on exploiting knowledge triplets~\cite{liu2018knowledge,zhu2017flexible} or knowledge graphs~\cite{liu2019knowledge,zhou2018commonsense,tuan2019dykgchat,huang2019knowledge,xu2020conversational}, the latter conditions on paragraph text~\cite{ghazvininejad2018knowledge,meng2020dukenet,kim2020sequential,lian2019learning}. 
For structured KGC, \citet{liu2018knowledge} utilize a neural knowledge diffusion module to encode knowledge triplets to predict related entities. \citet{liu2019knowledge} augment a knowledge graph to integrate with dialogue contexts in an open-domain dialogue. 
\citet{tuan2019dykgchat} assess a model's ability to reason multiple hops using a Markov chain over a constructed transition matrix, so that the model can zero-shot adapt to updated, unseen knowledge graphs. 
\citet{xu2020conversational} represent prior dialogue transition information as a knowledge graph and learn a graph grounded dialogue policy for generating coherent and controllable responses.
\citet{lei2020interactive} construct a user-item-attribute knowledge graph and ingeniously formalize dialogue policy learning as path reasoning on the graph.

Unlike most structured \ac{KGC} methods that select knowledge from open-domain knowledge-bases, \ac{MDG} aims to explore a multi-hop knowledge path transferred from patient states to physician actions using dedicated medical-domain knowledge graphs.

\negskip
\section{Method}
\label{sec:method}

\subsection{Problem formulation}
\label{sec:problem}

\OurParagraph{Medical dialogue systems.} 
Given $T$ dialogue turns, a medical dialogue session $d$ consists of a sequence of utterances, i.e., $d=\{U_{1}, R_1, U_2, R_2,\ldots,$ $U_{T}, R_T\}$, where $U_t$ and $R_t$ refers to utterances from a patient and responses from a virtual physician, respectively.
At the $t$-th turn, given the $t$-th patient utterance $U_t$ and previous physician response $R_{t-1}$, the dialogue system generates a response $R_t$.
Let $|U_t|$ be the number of words in $U_t$, we define $U_t = (U_{t,1}, U_{t, 2}, \ldots, U_{t, |U_t|})$ as a sequence of words.
The full vocabulary is defined as $\mathcal{V}$.
$K$ denotes an external knowledge base in the medical dialogue system, where each triplet in $K$ indicates a head entity, a relation, and a tail entity. 
Following~\cite{wang2019kgat}, we construct a knowledge graph $G^\mathit{global}$ by linking all triplets with overlapping entities (i.e., two triples will be linked iff they share overlapping entities) in $K$.
We assume that each entity is categorized into a set of entity types, i.e., $E_{type}={}$\{\emph{disease}, \emph{symptoms}, \emph{medicines}, \emph{treatments}\}.

We consider \OurMethod{} as a model with parameters $\theta$. Given the dialogue context, responses, and the knowledge graph $G^\mathit{global}$, we aim to maximize the probability distribution over $d$ in \OurMethod{}:
\begin{equation}
\small
    \prod_{t=1}^T p_{\theta}(R_t|R_{t-1}, U_{t}, G^\mathit{global}).
    \label{eq:rt1}
    \vspace{-2mm}
\end{equation}

\OurParagraph{Patient states and physician actions.} 
Text-span based dialogue state trackers have the double advantage of simplicity and good interpretability~\cite{wen2016network,lei2018sequicity,jin2018explicit}.
Hence, at the $t$-th turn, we define a text span $S_t$ (i.e., a sequence of words) as the \emph{patient state} to summarize past utterances and responses {(i.e., $U_{1}, R_{1}, \ldots, R_{t-1},U_t$)}.
Then we take $S_t$ as constraints to search in a knowledge base. 
Similar to $S_t$, we also use a text span $A_{t}$ to represent the \emph{physician action} at the $t$-th turn, which summarizes the physician's policy such as diagnose, medicine, or treatment.
$A_t$ is predicted through a policy learning process given $S_t$.
Thus, task completion in \ac{MDG} becomes a problem of generating two successive text spans, $S_t$ and $A_t$, at each turn.

\begin{figure}[!t]
    \centering
    \includegraphics[width=0.7\linewidth]{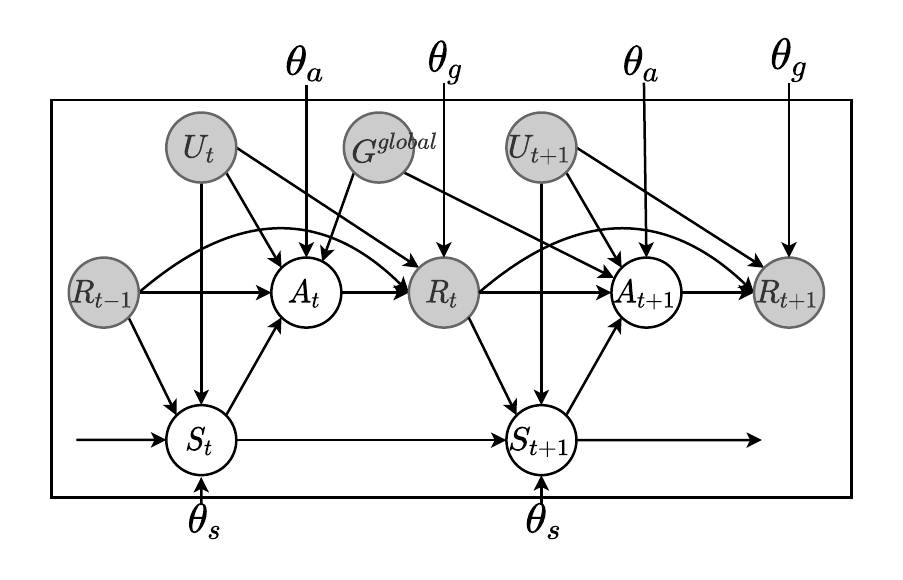}
    \caption{The graphical representation of \OurMethod. Shaded nodes represent observed variables.}
    \label{fig:directed-graph-of-vkbot}
\end{figure}

As text spans also help to improve the performance of response generation~\cite{lei2018sequicity,jin2018explicit}, generating $S_t$ and $A_t$ at each turn is a key component in \ac{MDG}.
In this paper, the problem of \ac{MDG} is decomposed into three successive steps: 
\begin{enumerate*}[label=(\arabic*)]
\item generating a state span $S_t$; 
\item generating an action span $A_t$; and 
\item generating the response $R_t$. 
\end{enumerate*}

\OurParagraph{Variational Bayesian generative model.} Large volumes of intermediate annotations for patient states and physician actions are impractical in \ac{MDG}. 
Thus, in \OurMethod{} we regard $S_t$ and $A_t$ as latent variables within a Bayesian generative model, so we reformulate Eq.~\ref{eq:rt1} as:
\begin{equation}
\small
\begin{split}
    \prod_{t=1}^{T} \sum_{S_t, A_t}  p_{\theta_g}(R_t|R_{t-1},U_t,S_t, A_{t}) 
    \cdot p_{\theta_{s}}(S_{t}) \cdot p_{\theta_a}(A_t),
\end{split}
\label{eq:vkbot-target}
\end{equation}
where $p_{\theta_g}(R_t|R_{t-1},U_t,S_t, A_{t})$ is derived using a \emph{response generator}, and $p_{\theta_s}(S_t)$ and $p_{\theta_a}(A_t)$ are estimated through a \emph{patient state tracker} and a \emph{physician policy network}, respectively. 

The graphical representation of \OurMethod{} is shown in Fig.~\ref{fig:directed-graph-of-vkbot}, where shaded and unshaded nodes indicate observed and latent variables, respectively. We see that a dependency exists between two adjacent states.
At $t$, $S_t$ is derived depending on previous state $S_{t-1}$, response $R_{t-1}$, and utterance $U_t$; subsequently, $A_t$ is inferred using $S_t$, $R_{t-1}$, $U_t$, and $G^\mathit{global}$. Thus, we calculate $p_{\theta_s}(S_t)$ and $p_{\theta_a}(A_t)$ as:
\begin{equation}
\label{eq:32}
\small
    \begin{split}
        \mbox{}\hspace*{-2mm}
        p_{\theta_s}(S_t) & \triangleq p_{\theta_{s}}(S_t|S_{t-1}, R_{t-1}, U_t) \text{ (\emph{prior state tracker})}, \\
        \mbox{}\hspace*{-2mm}
        p_{\theta_a}(A_t) & \triangleq p_{\theta_{a}}(A_t|S_{t}, R_{t-1}, U_t, G^{global}) \text{ (\emph{prior policy network})},
    \end{split}
\end{equation}

\begin{figure*}[!t]
  \centering
  \includegraphics[width=0.95\linewidth]{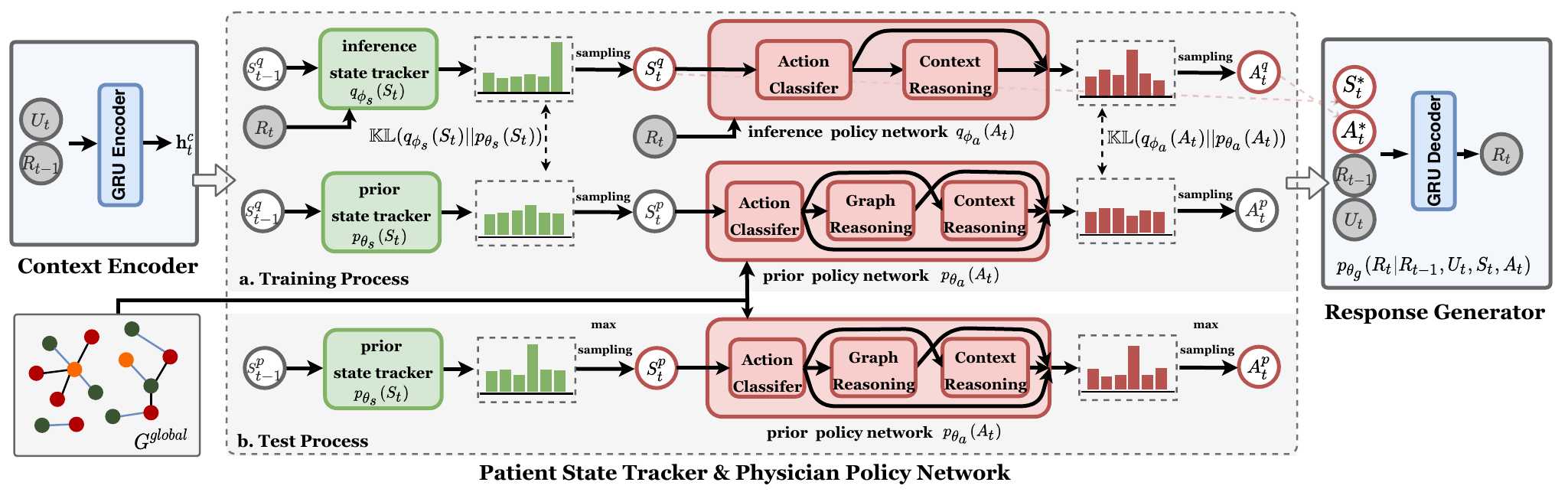}
  \caption{An overview of \OurMethod{}. We divide \OurMethod{} into a context encoder, a patient state tracker, a physician policy network, and a response generator. 
  Labels \textbf{a}, \textbf{b} indicate different sampling procedure in training and test process respectively.}
  \label{fig:model-ov}
\end{figure*}

\noindent where $\theta_s$ and $\theta_a$ are parameters; and a fixed initial value is assigned to $S_{0}$ at the beginning.
In \OurMethod{} we propose two \textit{prior networks} to estimate probabilistic distributions in Eq.~\ref{eq:32}, i.e., a prior state tracker and prior policy network.
Eventually, we draw a response $R_t$ from $p_{\theta_g}(R_t|R_{t-1},U_t,S_t, A_{t})$, with parameters $\theta_g$. 

To maximize Eq.~\ref{eq:vkbot-target}, we estimate the posterior distribution $p_{\theta}(S_t,$ $A_t|R_t, R_{t-1}, U_t, G^\mathit{global})$. 
However, the exact posterior distribution is intractable due to its complicated posterior expectation estimation. 
To address this problem, we introduce two \emph{inference networks}~\cite{kingma2013auto} (i.e., $q_{\phi_s}(S_t)$ and $q_{\phi_a}(A_t)$) to approximate the posterior distributions over $S_t$ and $A_t$, respectively: 
\begin{equation}
\small
    \begin{split}
        \mbox{}\hspace*{-2mm}
        q_{\phi_s}(S_t) &\triangleq q_{\phi_s}(S_t|S_{t-1}, R_{t-1}, U_t, R_t) \text{ (\emph{inference state tracker})}, \\
        \mbox{}\hspace*{-2mm}
        q_{\phi_a}(A_t) &\triangleq q_{\phi_a}(A_t|S_t, R_{t-1}, U_t, R_t) \text{ (\emph{inference policy network})},
    \end{split}
\end{equation}
where $\phi_s$ and $\phi_a$ are parameters in inference networks.

\OurParagraph{Evidence lower bound (ELBO).} At $t$, we derive the ELBO to optimize both prior and inference networks simultaneously as follows:   
\begin{equation}
\small
\label{eq: joint-elbo-optim}
    \begin{split}
         & \log p_{\theta}(R_t|R_{t-1}, U_t, G^\mathit{global})  \\ 
         & \ge \mathbb{E}_{q_{\phi_s}(S_{t-1})} \Big[ \mathbb{E}_{q_{\phi_{s}}(S_{t}) \cdot q_{\phi_a}(A_t)} [ \log p_{\theta_g}(R_{t}|R_{t-1}, U_t, S_t, A_t)] \\  
        & \quad - \mathbb{KL}(q_{\phi_{s}}(S_{t}) || p_{\theta_{s}}(S_t)) - \mathbb{KL}( q_{\phi_a}(A_{t}) \|  p_{\theta_a}({A_t}))\Big] \\
        & = - \mathcal{L}_\mathit{joint},
    \end{split}
\end{equation}
where $\mathbb{E}({\cdot})$ is the expectation, and $\mathbb{KL}(\cdot\|\cdot)$ denotes the Kullback-Leibler divergence. 
To estimate Eq.~\ref{eq: joint-elbo-optim}, from $q_{\phi_s}(S_{t-1})$ we first draw a state $S^q_{t-1}$, which is for estimating $p_{\theta_s}(S_t)$ and $q_{\phi_s}(S_t)$; then, $S^p_t$ is drawn from $p_{\theta_s}(S_t)$ and $S^q_t$ is obtained through $q_{\phi_s}(S_t)$. We estimate $p_{\theta_a}(A_t)$ and $q_{\phi_a}(A_t)$ using $S^p_t$ and $S^q_t$, respectively, and draw $A^q_t$ from $q_{\phi_a}(A_t)$. Finally, $p_{\theta_g}(R_t|\cdot)$ generates $R_t$ depending on $S^q_t$ and $A^q_t$. The above sampling procedure is shown in Fig.~\ref{fig:model-ov} (a. Training process).

\negskip
\subsection{Context encoder}
At $t$, we encode the dialogue history $(R_{t-1}, U_t)$ into a list of word-level hidden vectors $\boldsymbol{H}_t=(\boldsymbol{h}_{t,1}, \ldots, \boldsymbol{h}_{t,|R_{t-1}|+|U_t|})$ using a bi-directional Gated Recurrent Unit  (GRU)~\cite{cho2014properties}:
\begin{equation}
\label{eq:his_encode}
\small
\begin{split}
	\boldsymbol{H}_{t} & = \text{BiGRU}(\boldsymbol{h}_{t-1}^{c}, \boldsymbol{e}^{R_{t-1}}_1, \boldsymbol{e}^{R_{t-1}}_2, \ldots,\boldsymbol{e}^{R_{t-1}}_{|R_{t-1}|},\ldots, \boldsymbol{e}^{U_{t}}_{|U_t|}),
\end{split}
\end{equation}
where $|R_{t-1}|$ and $|U_t|$ indicate the number of words in $R_{t-1}$ and $U_t$ respectively;  $\boldsymbol{e}^{R_{t-1}}_i$ denotes the embedding of the $i$-th word in $R_{t-1}$. Initializing from the hidden representation $\boldsymbol{h}^c_{t-1}$ of the $(t-1)$-th turn, the last hidden state $\boldsymbol{h}_{t,|R_{t-1}|+|U_t|}$ attentively read $\boldsymbol{H}_t$ to get the $t$-th turn's hidden representation, i.e.,  $\boldsymbol{h}^c_{t}$.

\negskip
\subsection{Patient state tracker}
\label{sec:model-st}

As we formulate patient states as text spans, the prior and inference state trackers are both based on an encoder-decoder framework.
We encode $S^q_{t-1}$ using a GRU encoder to get $\boldsymbol{h}^{S^{q}}_{t-1}$ during the encoding procedure. 
We then incorporate $\boldsymbol{h}^{S^{q}}_{t-1}$ with $\boldsymbol{h}^{c}_t$ to infer the prior state distribution $p_{\theta_s}(S_{t})$ at the $t$-th turn. 
During the decoding procedure, we first infer the prior distribution over the patient state.
We denote $\boldsymbol{b}^{S^{p}}_{t,0} = \boldsymbol{W}_s^p[\boldsymbol{h}_{t}^c ; \boldsymbol{h}_{t-1}^{S^{q}}]$ as the initial hidden representation of the decoder, where $\boldsymbol{W}_s^p$ is a learnable parameter matrix, and $[\cdot ;\cdot ]$ denotes vector concatenation. 
At the $i$-th token during decoding, the decoder sequentially decodes $S_{t}$ to output $\boldsymbol{b}^{S^{p}}_{t,i}$ given previous token embedding $\boldsymbol{e}^{S^{p}}_{t,i-1}$, 
next projects $\boldsymbol{b}^{S^{p}}_{t,i}$ into the patient state space. We set $S_t$'s length to |S|, and the prior distribution over $S_t$ is calculated as:
\begin{equation}
\small
     p_{\theta_s}(S_t) = \prod_{i=1}^{|S|}{\softmax(\text{MLP}(\boldsymbol{b}^{S^{p}}_{t, i}))},
\end{equation}
where $\text{MLP}$ is a multilayer perceptron (MLP)~\cite{gardner1998artificial}. 
To approximate the state posterior distribution, the inference state tracker follows a similar process but additionally incorporates the encoding of $R_t$, i.e., $\boldsymbol{h}^{R}_{t}$. 
The GRU decoder is initialized as $\boldsymbol{b}^{S^{q}}_{t,0} = \boldsymbol{W}^q_s[\boldsymbol{h}^c_t ; \boldsymbol{h}^{S^{q}}_{t-1} ; \boldsymbol{h}^{R}_t]$, where $\boldsymbol{W}^q_s$ is a learnable parameter, and it outputs $\boldsymbol{b}^{S^q}_{t,i}$ at the $i$-th decoding step. 
Accordingly, we write the approximate posterior distribution as:
\begin{equation}
\small
    q_{\phi_s}(S_t) = \prod_{i=1}^{|S|}{\softmax(\text{MLP}(\boldsymbol{b}^{S^{q}}_{t,i}))}.
    \vspace{-2mm}
\end{equation}

\negskip
\subsection{Physician policy network}
\label{sec:ppn}
The prior and inference policy networks are also based on an encoder-decoder structure. 
Specifically, we represent $A_{t}$ as a pair of an action category $A^c_{t}$ and a list of explicit keywords $A^k_{t}$, i.e., $A_t=\{A^c_{t}, A^k_{t}\}$. Here we set the length of $A^k_t$ to |A|.

As for the prior policy network, at the beginning of the encoding procedure, we encode $S^{p}_t$ to a vector $\boldsymbol{h}^{S^p}_t$ using a GRU encoder. 
Furthermore, external knowledge is important for the physician network to react given the patient state.
As the external medical knowledge graph $G^\mathit{global}$ is large (in the number of entities), we extract a sub-graph $G^\mathit{local}_n$ from $G^\mathit{global}$ via a knowledge base retrieval operation \textit{\textbf{qsub}},
where we regard each entity in $S^p_{t}$ as seed nodes during \textit{\textbf{qsub}}. Starting from $S^p_{t}$, we extract all the accessible nodes and edges in $G^\mathit{global}$ within $n$ hops to get the sub-graph $G^\mathit{local}_n$~\cite{wang2018ripplenet}. 
Besides, we link all the entities appear in $S^p_t$ to ensure $G^\mathit{local}_n$ is connected.

To combine the relation type during information propagation, we employ the relational graph attention network (RGAT)~\cite{busbridge2019relational} to represent each entity in the external knowledge graph. 
Given a graph $G =\{X,Y\}$ including relations $Y$ and nodes $X$, after multiple rounds of propagation, RGAT outputs a feature matrix $\boldsymbol{G} = [\boldsymbol{g}_1, \boldsymbol{g}_2, \ldots, \boldsymbol{g}_X]$, where $\boldsymbol{g}_x$ ($1\le x \le X$) is the embedding of node $x$. We use $\text{RGAT}$ to denote this operation, so we have: $\boldsymbol{G}^\mathit{local}_n=\text{RGAT}(G^\mathit{local}_n)$.

To decode outputs, we infer $A^c_{t}$ and $A^k_{t}$ sequentially.
We devise an action classifier to infer $A^c_{t}$. 
Following~\cite{bahdanau2014neural}, we compute an attention vector $\boldsymbol{q}_t$ over $\boldsymbol{G}^\mathit{local}_n$ with $\boldsymbol{h}^c_t$ as the query. Sequentially, the action classifier incorporates $\boldsymbol{q}_t$, and classifies physician action into four categories, i.e., \textit{ask symptoms}, \textit{diagnosis}, \textit{prescribe medicine} and \textit{chitchat}, as follows:
\begin{equation}
\label{eq:action-classifier-fn}
\small
        P_{\theta_{a,c}}(A^c_t) = 
        \softmax(\boldsymbol{W}^p_c[
        \boldsymbol{h}^{S^{p}}_{t} ; 
        \boldsymbol{h}^{c}_{t} ; 
        \boldsymbol{q}_t]
        ),
\end{equation}
where $\boldsymbol{W}^p_c$ is a learnable parameter. Then we  draw an action category $A^{c,p}_{t}$ by sampling from $p_{\theta_{a,c}}(A^c_t)$. 

$A^k_{t}$ is decoded sequentially based on a GRU decoder. 
To infer the prior probabilistic distribution, two reasoning detectors (i.e., a context-reasoning detector and a graph-reasoning detector) are proposed to corporately project the hidden representation of the decoder to the action space at each decoding step. 
The decoder is initializes as $\boldsymbol{b}^{A^{k,p}}_{t,0} = \boldsymbol{W}^p_k[ \boldsymbol{h}^{S^p}_t ; \boldsymbol{h}^c_t ; \boldsymbol{e}^{A^{c,p}}_{t}]$,  
where $\boldsymbol{e}^{A^{c,p}}_{t}$ is the embedding of $A^{c,p}_{t}$. At the $i$-th decoding step, 
the decoder outputs $\boldsymbol{b}^{A^{k,p}}_{t,i}$.
The context-reasoning detector and the graph-reasoning detector together infer $A^{k}_{t,i}$ with $\boldsymbol{b}^{A^{k,p}}_{t,i}$.

Learning from the raw context and state, the context-reasoning detector infers the prior distribution over $A^k_{t,i}$ using a MLP as follows:
\begin{equation}
\small
    \begin{split}
        p_{\theta_{a,d}}(A^k_{t,i}) = \frac{1}{z_A} \exp{( {\text{MLP}([\boldsymbol{h}^{S^{p}}_t ; \boldsymbol{h}^{c}_t ; \boldsymbol{b}^{A^{k,p}}_{t,i}])})},
    \end{split}
\end{equation}
where $z_A$ is the normalization term shared with the graph-reasoning detector. The graph-reasoning detector considers to copy entities from $G^\mathit{local}_n$:
\begin{equation}
\small
    p_{\theta_{a,g}}(A^{k}_{t,i}) =
    \frac{1}{z_A} \mathbb{I}(e_j, A^{k}_{t,i}) \cdot
    \exp{(\boldsymbol{W}_g[\boldsymbol{h}^{c}_t; \boldsymbol{b}^{A^{k,p}}_{t,i}; \boldsymbol{g}_j])},
\end{equation}
where $\boldsymbol{W}_g$ is a learnable parameter matrix, $e_j$ is the $j$-th entity in $G^\mathit{local}_n$, $\boldsymbol{g}_j$ is the $j$-th entry embedding of $\boldsymbol{G}^\mathit{local}_n$, $\mathbb{I}(e_j, A^{k}_{t,i})$ equals 1 if $e_j = A^k_{t,i}$ otherwise 0. 
Eventually, we calculate the prior distribution over $A_{t}$ as follows:
\begin{equation}
\small
    \begin{split}
        p_{\theta_a}(A_t) = p_{\theta_{a,c}}(A^c_{t}) \cdot \prod_{i=1}^{|A|}{[p_{\theta_{a,d}}(A^k_{t,i}) + p_{\theta_{a,g}}(A^k_{t,i})]}.
    \end{split}
\end{equation}
The inference policy network approximates the action category posterior distribution and keywords posterior distribution by extracting indicative information from the response $R_t$. A GRU encoder encodes $R_t$ to $\boldsymbol{h}^R_t$, $S^{q}_t$ to $\boldsymbol{h}^{S^q}_t$ respectively.
Then we get the action category approximate posterior distribution as follows:
\begin{equation}
\small
        q_{\phi_{a, c}}(A^c_{t}) =  \softmax(\boldsymbol{W}^q_c[
        \boldsymbol{h}^{c}_t ; 
        \boldsymbol{h}^{S^{q}}_t;
        \boldsymbol{h}^{R}_t
        ]).
\end{equation}
Thereafter, we draw $A^{c,q}_{t}$ via sampling from $q_{\phi_{a,c}}(A^{c}_{t})$.
To reinforce the effect of information from $R_t$, we only use the context-reasoning detector to approximate the posterior distribution of $A^k_t$. 
The decoder is initialized as $\boldsymbol{b}^{A^{k,q}}_{t,0} = \boldsymbol{W}^q_k[\boldsymbol{h}^c_t ; \boldsymbol{h}_t^{S^{q}} ; \boldsymbol{e}^{A^{c,q}}_{t} ; \boldsymbol{h}^{R}_t]$, 
where $\boldsymbol{e}^{A^{c,q}}_{t}$ is the embedding of $A^{c,q}_{t}$, $\boldsymbol{W}^q_k$ reflects a learnable parameter matrix. 
At the $i$-th decoding step, the decoder outputs $\boldsymbol{b}^{A^{k,q}}_{t,i}$, so we have the approximate posterior distribution over the $i$-th action keyword:
\begin{equation}
\small
    q_{\phi_{a,d}}(A^k_{t,i}) = \softmax (\text{MLP}([
    \boldsymbol{h}^{c}_t ;
    \boldsymbol{h}^{S^{q}}_t ; 
    \boldsymbol{b}^{A^{k,q}}_{t,i}])).
\end{equation}
Eventually, we get the approximate posterior distribution of $A_t$: 
\begin{equation}
\small
    \begin{split}
        q_{\phi_a}(A_{t}) & = q_{\phi_{a,c}}(A^{c}_{t}) \cdot \prod_{i=1}^{|A|} q_{\phi_{a,d}}(A^{k}_{t,i}).
    \end{split}
\end{equation}
Inspired by \citet{jin2018explicit}, we also employ the copy mechanism in $p_{\theta_s}(S_t)$ and $q_{\phi_s}(S_t)$, so as to copy tokens from $R_{t-1}, U_t, S^q_{t-1}$. In the same way, we copy tokens from $R_t$ for $q_{\phi_{a}}(A_{t})$.

\negskip
\subsection{Response generator} 
At the first stage during the response generation, we use a GRU encoder to encode $S^q_{t}$ into $\boldsymbol{S}^q_t$ which is a word-level embedding matrix of $S^q_t$. Each column vector in $\boldsymbol{S}^q_t$ reflects an embedding vector of the corresponding word in $S^q_t$.
In the same manner, $A^{k,q}_{t}$ is encoded to $\boldsymbol{A}^{k,q}_{t}$. 
As mentioned in Sec.~\ref{sec:model-st} and~\ref{sec:ppn}, we also calculate the holistic embedding $\boldsymbol{h}^{S^{q}}_t$ and $\boldsymbol{h}^{A^{k,q}}_{t}$ from $S^q_{t}$ and $A^{k,q}_{t}$, respectively. 
The response decoder with a GRU cell takes $\boldsymbol{b}^R_{t, 0} = \boldsymbol{W}_d[ \boldsymbol{h}^{c}_t; \boldsymbol{h}^{S^{q}}_t; \boldsymbol{e}^{A^{c,q}}_{t}; \boldsymbol{h}^{A^{k,q}}_{t}]$ as the initial hidden state.

At the $i$-th decoding step, the output $\boldsymbol{b}^{R}_{t,i-1}$ from the $i-1$-th step attentively reads the context representation $\boldsymbol{H}_t$ to get $\boldsymbol{b}_{t, i}^{h}$,
Meanwhile, $\boldsymbol{b}^{R}_{t,i-1}$ attentively read $\boldsymbol{S}^{q}_{t} and \boldsymbol{A}^{k,q}_{t}$ to get $\boldsymbol{b}^{s}_{t, i}$ and $\boldsymbol{b}^{a}_{t, i}$ respectively. 
Subsequently, 
$[\boldsymbol{b}_{t, i}^{h}; \boldsymbol{b}_{t, i}^{s}; \boldsymbol{b}_{t, i}^{a}; \boldsymbol{e}^R_{t, i-1}]$ 
are fed into the decoder GRU cell to output $\boldsymbol{b}^R_{t, i}$, where $\boldsymbol{e}^R_{t, i-1}$ is the embedding of $(i-1)$-st word in $R_{t}$. The probability of generating $R_{t,i}$ is formulated as a sum of the generative probability and a copy term:
\begin{equation}
\small
    \begin{split}
        p_{\theta_g}(R_{t,i}) & =
        p_{\theta_g}^g(R_{t,i}) +  p_{\theta_g}^c(R_{t,i}),\\
        p_{\theta_g}^g(R_{t,i}) & = \frac{1}{z_R} \exp{(\text{MLP}(\boldsymbol{b}^{R}_{t,i}))},\\
        p_{\theta_g}^c(R_{t,i}) & = 
        \frac{1}{z_R} \sum_{j:W_j = R_{t,i}} \exp{({\boldsymbol{h}^{W}_j}^{\mathsf{T}} \cdot \boldsymbol{b}^R_{t,i})},
    \end{split}
\end{equation}
%
\noindent where $p_{\theta_g}^g(R_{t,i})$ is the generative probability, $p_{\theta_g}^c(R_{t,i})$ is the copy term, $z_R$ is the normalization term shared with $p_{\theta_g}^c(R_{t,i})$. 
We write $W$ for a concatenation sequence of $R_{t-1}$, $U_t$, $ S^q_{t}$, and $A^{k,q}_t$, where $W_j$ is the $j$-th word in $W$,  and $\boldsymbol{h}^{W}_j$ is the $j$-th vector in $[\boldsymbol{H}_t; \boldsymbol{S}^q_t; \boldsymbol{A}^{k,q}_{t}]$.

\subsection{Collapsed inference and training}
\label{sec:variational-inference}
\begin{figure}
  \centering
  \includegraphics[width=0.92\linewidth]{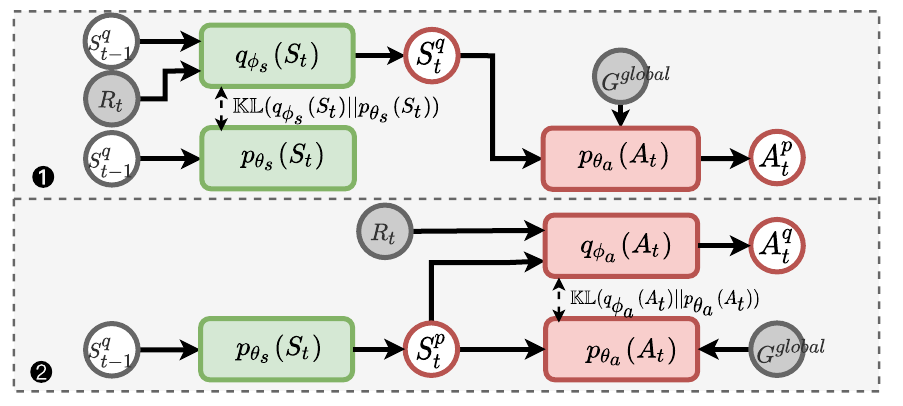}
  \caption{The graphical representation of 2-stage collapsed inference.}
  \label{fig:2s-collapsed-inference}
\end{figure}
Eq.~\ref{eq: joint-elbo-optim} provides a unified objective for optimizing all components.
However, the joint distribution $p_{\theta_{s}}(S_t) \cdot p_{\theta_a}(A_t)$ is hard to optimize as $p_{\theta}(A_t)$ is easily misled with incorrect sampling results of $S^p_t$ from $p_{\theta_s}(S_t)$.
To address this problem, we propose a \emph{2-stage collapsed inference} method by decomposing the objective function into 2-stage optimization objectives. 
During the first stage, 
we fit $p_{\theta_s}(S_t)$ to $q_{\phi_s}(S_t)$ to derive the ELBO (labeled by \ding{202} in Fig.~\ref{fig:2s-collapsed-inference}):
\begin{equation}
 \small
    \begin{split}    
    & \mbox{}\hspace*{-2mm}\log p_{\theta}(R_t| R_{t-1}, U_t, G^\mathit{global}) \\ 
    & \mbox{}\hspace*{-2mm}\ge \mathbb{E}_{q_{\phi_s}(S_{t-1})} 
    \Big[\mathbb{E}_{q_{\phi_s}(S_t)} \big[\mathbb{E}_{p_{\theta_a}({A_t
    })} [ \log p_{\theta_g}(R_t|R_{t-1}, U_t, S_t, A_t)]\big] 
        \hspace*{-2mm}\mbox{}\\ 
    & \mbox{}\hspace*{-2mm}\quad - \mathbb{KL}(q_{\phi_s}(S_t)\|p_{\theta_s}(S_t)) \Big] \\
    & \mbox{}\hspace*{-2mm}= - \mathcal{L}_{s}.
    \label{eq:vi-st}
    \end{split}
\end{equation}
Subsequently, similar to the optimization of $\phi_s$ and $\theta_s$, $p_{\theta_a}(A_t)$ is fit $q_{\phi_a}(A_t)$ to formulate the ELBO (labeled by \ding{203} in Fig.~\ref{fig:2s-collapsed-inference}) as follows:
\begin{equation}
 \small
\begin{split}
    & \mbox{}\hspace*{-2mm}\log p_{\theta}(R_t| R_{t-1}, U_t, G^\mathit{global}) \\ 
    & \mbox{}\hspace*{-2mm}\ge  \mathbb{E}_{q_{\phi_s}(S_{t - 1})} \Big[ \mathbb{E}_{p_{\theta_s}(S_t)} \big[ \mathbb{E}_{q_{\phi_a}(A_t)} [ \log p_{\theta_g}(R_t|R_{t-1}, U_t, S_t, A_t)] 
        \hspace*{-2mm}\mbox{}\\ 
    &\mbox{}\hspace*{-2mm}\quad - \mathbb{KL}(q_{\phi_a}(A_t)\|p_{\theta_a}(A_t))\big]\Big] \\
    & \mbox{}\hspace*{-2mm}= - \mathcal{L}_a.
\end{split}
\label{eq:vi-at}
\end{equation}
%
Accordingly, the training procedure comprises two stages when no human-annotation exist. So we have:
\begin{equation}
\small
\mathcal{L}^{un} = \left\{
\begin{array}{ll}
\mathcal{L}_s & \text{(1st training stage)} \\
\mathcal{L}_s + \mathcal{L}_a & \text{(2nd training stage)}. \\
\end{array}
\right.
\end{equation}
\noindent We first minimize $\mathcal{L}_s$ to get proper state tracking results. 
Then we jointly train all parameters to the 2nd stage optimization. 
We learn \OurMethod{} by SGVB and draw samples with the Gumbel-Softmax trick \cite{jang2016categorical} to calculate the gradients with discrete variables. 

If annotated states $\bar{S}_t$ and actions $\bar{A}_t$ are partially available, we add the auxiliary loss $\mathcal{L}^{sup}$ to perform semi-supervised training:
\begin{equation}
\small
\begin{split}
    \mathcal{L}^{sup} = & - ( 
    \log p_{\theta_g}(R_{t}|\bar{S}_t, \bar{A}_t, R_{t-1}, U_t) \\ 
    & + \log (p_{\theta_s}(\bar{S}_t) \cdot q_{\phi_s}(\bar{S}_{t})) 
    + \log( p_{\theta_a}(\bar{A}_t) \cdot q_{\phi_a}(\bar{A}_t))).
\end{split}
\end{equation}
In the test process, we only execute $p_{\theta_s}(S_t)$ and $p_{\theta_a}(A_t)$ to infer patient states and physician actions (labeled by \textbf{b} in Fig.~\ref{fig:model-ov}).

\negskip
\section{Experimental Setup}
\label{sec:exp}
\smallnegskip
\subsection{Research questions}
\smallnegskip
We seek to answer the following research questions: 
\begin{enumerate*}[label=(RQ\arabic*)]
\item How does \OurMethod{} perform on medical dialogue generation? Is unlabeled data helpful for generating accurate responses? 
\item What is the effect of each component in \OurMethod{}? Are the reasoning detectors helpful to improve physician action prediction?
\item What is the effect of the length of the patient state and physician action in \OurMethod{}? 
\item Can \OurMethod{} provide interpretable results?
\end{enumerate*}

\negskip
\subsection{Datasets}
We adopt three medical dialogue datasets for our experiments, all of which are collected from real-world medical consultation websites after data anonymization, i.e., close to clinically authentic medical scenarios.
Two have been applied in previous studies, and we propose a new dataset with large-scale external knowledge. 

Existing medical dialogue datasets have a limited amount of external knowledge, a limited length of dialogues, and a handful of medical departments. These constraints make it difficult to evaluate \ac{MDG} approaches. 
To address this problem, we collect a large-scale dataset \textbf{K}nowledge-\textbf{a}ware \textbf{Med}ical conversation dataset (\textbf{KaMed}) from ChunyuDoctor,\footnote{\url{https://www.chunyuyisheng.com/}} a large online Chinese medical consultation platform. 
The dataset caters for challenging and diverse scenarios, as it contains over $100$ hospital departments with a large-scale external knowledge graph. 
To simulate realistic clinical conversational scenarios, in KaMed the average number of rounds of a dialogue is $11.62$, much longer than existing medical dialogue datasets. 
Unlike other medical dialogue datasets, KaMed is equipped with large-scale external medical knowledge,  crawled from CMeKG,\footnote{\url{http://zstp.pcl.ac.cn:8002}} the largest Chinese medical knowledge platform.

To evaluate \OurMethod{}, we also use two benchmark datasets.
MedDG \cite{liu2020meddg} is collected from ChunyuDoctor, related to 12 types of common gastrointestinal diseases, and provides semi-automatic annotated states and actions; the average number of rounds of a dialogue session is $9.92$. 
MedDialog~\cite{chen2020meddiag} is collected from an online medical platform. We filter out dialogues with fewer than three rounds, but the average number of rounds is still relatively low, only $4.76$.
We also collect relevant medical knowledge for the MedDG and MedDialog datasets. 
The dataset statistics are listed in Table~\ref{tab:statistic-of-datasets}.

\negskip
\subsection{Baselines and comparisons}
In the context of RQ1, we write \OurMethod{}$\backslash$un for the model that is only trained using annotated data. We devise a variation of \OurMethod{} by replacing the GRU encoder with Bert, and use \OurMethod{}-Bert to denote it.
In the context of RQ2, we write \OurMethod{}$\backslash$S for the model that eliminates the latent state variable, \OurMethod{}$\backslash$A for the model that eliminates the latent action variable, \OurMethod{}$\backslash$G for the model without the graph-reasoning detector, \OurMethod{}$\backslash$C for the model without the context-reasoning detector, and \OurMethod{}$\backslash$2s for the model without 2-stage collapsed inference (i.e., minimizing $\mathcal{L}_\mathit{joint}$ in Eq.~\ref{eq: joint-elbo-optim}). 

As far as we know, only \citet{liu2020meddg} have addressed the same task as we do. Thus, for \ac{MDG}, we use HRED-Bert~\cite{liu2020meddg} as a baseline, which integrates Bert~\cite{devlin2018bert} with the HRED model for \ac{MRG}. 
We consider three types of baseline: open-domain dialogue generation, knowledge grounded conversations, and task-oriented dialogue generation. 
As open-domain approaches, we use Seq2Seq~\cite{sutskever2014sequence}, HRED~\cite{serban2016building}, and VHRED~\cite{serban2016hierarchical} as baselines. 
Seq2Seq is a sequence-to-sequence generation model with attention and copy mechanism~\cite{gu2016incorporating}; 
HRED uses a hierarchical encoder-decoder structure to model the dialogue at the word- and utterance-level; VHRED extends HRED with a continuous latent variable to facilitate generation.
As knowledge-grounded methods, we use CCM~\cite{zhou2018commonsense}, NKD~\cite{liu2018knowledge}, and PostKS~\cite{lian2019learning} as baselines. 
CCM applies two graph attention mechanisms to augment the semantic information during response generation; NKD uses a neural knowledge diffusion module to retrieve relevant knowledge; 
PostKS uses dialogue context and responses to infer the posterior knowledge distribution.  
For task-oriented dialogue generation, we use SEDST~\cite{jin2018explicit}, LABES~\cite{zhang2020probabilistic}, MOSS~\cite{liang2020moss}, and DAMD~\cite{zhang2020task} as baselines. 
SEDST formalizes the dialogue state as a text-span to copy keywords from question to state; LABES regards the state as discrete latent variables to conduct the Straight-Through Estimator for calculating the gradient; MOSS incorporates supervision from various intermediate dialogue system modules; DAMD uses GRU-based decoders to decode the state, action, and response in a supervised manner. Similar performance can be also observed in Transformer based methods \cite{yang2020generation,zeng2020multi}.

\begin{table}
    \centering
    \small
    \caption{Statistics of KaMed, MedDialog and MedDG; \ding{51} under `An' indicates the dataset provides annotations.}
    \scalebox{0.95}{
    \begin{tabular}{l l l r c}
    \toprule
    \textbf{Dataset} & \textbf{Train/Valid/Test} & \textbf{Entity/Triplet} & \textbf{Turn} & \textbf{An} \\
    \midrule
    KaMed & $57,754/3,000/3,000$ & $5,682/53,159$ & $11.62$ & \ding{55} \\
    MedDialog & $32,723/3,000/3,000$ & $4,480/79,869$ & $4.76$ & \ding{55} \\
    MedDG & $14,864/2,000/1,000$ & $160/1,240$ & $9.92$ & \ding{51} \\
    \bottomrule
    \end{tabular}
    }
    \label{tab:statistic-of-datasets}
\end{table}

\begin{table*}
    \centering
    \small
    \caption{Automatic evaluation on the KaMed and MedDialog datasets. Boldface scores indicate best results, significant improvements over the best baseline are marked with * (t-test, $p<0.05$).}
    \scalebox{1.0}{
    \begin{tabular}{l l c c c c c c c c c c c c}
    \toprule
        Dataset & Model & B@2 & R@2 & D@1 & D@2 & ma-P & ma-R & ma-F1 & mi-P & mi-R & mi-F1 & EA & EG \\ 
        \midrule
        \multirow{7}*{KaMed} & Seq2Seq & 2.71 & 1.58 & 1.24 & 6.85 & 24.82 & 11.14 & 15.38 & 27.60 & 12.78 & 17.47 & 27.93 & 27.75 \\
        ~ & HRED	& 2.59 & 1.59 & 1.17 & 6.65 & 27.14 & 11.28 & 15.94 & 28.36 & 12.82 & 17.65 & 27.79 & 27.75 \\
        ~ & VHRED	& 2.49 & 1.55 & 1.15 & 6.42 & \textbf{28.65} & 11.18 & 16.08 & 28.36 & 12.61 & 17.46 & 27.44 & 27.36 \\
        ~ & CCM	& 2.42 & 1.47 & 0.51 & 2.12 & 19.05 & 10.59 & 13.61 & 23.53 & 14.49 & 17.93 & 33.64 & 33.44 \\
        ~ & SEDST	& 2.40 & 1.39 & 0.43 & 2.19 & 26.45 & 10.92 & 15.46 & \textbf{28.86} & 13.81 & 18.68 & 31.85 & 31.64 \\
        ~ & PostKS & 2.52 & 1.44 & 1.00 & 5.55 & 25.18 & 11.34 & 15.64 & 26.01 & 14.15 & 18.33 & 29.76 & 29.61  \\
        ~ & \OurMethod{}	& \textbf{2.89} & \textbf{1.97}\rlap{*} & \textbf{1.61} & \textbf{9.30} & 22.91 & \textbf{17.00}\rlap{*} & \textbf{19.52}\rlap{*} & 26.35 & \textbf{18.84}\rlap{*} & \textbf{21.97}\rlap{*} & \textbf{43.18}\rlap{*} & \textbf{43.11}\rlap{*} \\
        \midrule
        \multirow{7}*{MedDialog}
         & Seq2Seq	& 3.13 & 1.11 & 1.62 & 8.11 & 23.73 & 8.54 & 12.56 & 25.65 & 8.97 & 13.29 & 20.53 & 21.95 \\
        ~ & HRED	& 2.56 & 0.85 & 1.72 & 8.54 & 23.38 & 8.77 & 12.75 & 25.34 & 8.79 & 13.06 & 20.64 & 21.97 \\
        ~ & VHRED	& 2.82 & 1.01 & 1.74 & 8.84 & \textbf{26.23} & 8.87 & 13.26 & \textbf{26.29} & 9.00 & 13.41 & 20.15 & 21.43 \\
        ~ & CCM	& 3.29 & 1.14 & 1.42 & 6.91 & 20.36 & 9.49 & 12.94 & 20.68 & 10.82 & 14.21 & 26.51 & 27.81 \\
        ~ & SEDST	& 2.37 & 0.89 & 0.72 & 3.13 & 22.82 & 8.00 & 11.85 & 24.81 & 8.06 & 12.17 & 20.13 & 21.23 \\
        ~ & PostKS & 3.26 & 1.27 & 1.63 & 8.48 & 25.53 & 9.81 & 14.17 & 21.60 & 10.15 & 13.81 & 24.38 & 25.76 \\
        ~ & \OurMethod{}	& \textbf{3.96}\rlap{*} & \textbf{1.45}\rlap{*} & \textbf{2.14} & \textbf{12.18} & 22.77 & \textbf{14.11}\rlap{*} & \textbf{17.42}\rlap{*} & 23.50 & \textbf{14.73}\rlap{*} & \textbf{18.11}\rlap{*} & \textbf{34.51}\rlap{*} & \textbf{36.79}\rlap{*} \\
        \bottomrule
    \end{tabular}
    }
    \label{tab:automatic-evaluation-unsupervised}
\end{table*}
%

\negskip
\subsection{Evaluation metrics}
\textbf{Automatic evaluation.} 
To assess the language quality for the generated responses, we employ classical word-overlap based metrics, \textit{BLEU-2} (B@2)~\cite{papineni2002bleu} and \textit{ROUGE-2} (R@2)\cite{lin2004rouge}, to measure performance. As shortcomings have been reported for using BLEU/ROUGE to measure dialogue generation~\cite{liulsncp16}, we also use \textit{Distinct-1} (D@1) and \textit{Distinct-2} (D@2)~\cite{li2015diversity}, where Distinct-n is defined as the proportion of distinct n-grams in generated responses.
To measure the correctness of prediction results from the physician policy network, following \cite{liu2020meddg}, we calculate \textit{Precision} (P), \textit{Recall} (R), and \textit{F1} (F1) scores of predicted entities in the responses.
We adopt the prefix \textit{ma-} and \textit{mi-} to indicate macro-average and micro-average Precision, Recall, and F1 scores, respectively. 
We also employ embedding-based topic similarity metrics~\cite{liu2016not}, i.e., \textit{Embedding Average} (EA) and \textit{Embedding greedy} (EG), to evaluate the semantic relevance of the predicted entities between generated response and target response. We use mean explainability precision (MEP) and mean explainability recall (MER) to evaluate explainability~\citep{zhang2018explainable}.

\OurParagraph{Human evaluation.} 
We randomly sample 600 dialogues and their corresponding generations from our model as well as the baselines.
We recruit three professional annotators from a third-party hospital to evaluate the responses generated by different models.
Following~\citet{liu2020meddg}, we evaluate the responses generated by all models in terms of following three metrics: \textit{sentence fluency} (Flu), \textit{knowledge correctness} (KC), and \textit{entire quality} (EQ). 
Flu measures if the generated response is smooth; 
KC evaluates whether the response is correct; 
EQ measures the annotator's satisfaction with the generated response.
Three annotators are asked to rate each generated response with a score range from 1 (bad) to 5 (excellent) for each entry. Model names were masked out during evaluation.

\negskip
\subsection{Implementation details}
We conduct our experiments with a batch size of 16, and the size of embedding and the GRU hidden state set to $300$ and $512$, respectively. 
We set $n=2$ in the $\boldsymbol{qsub}$ operation.
The graph hidden size and output size are set to $128$ and $512$, respectively. All modules are trained in an end-to-end paradigm.
We use the pkuseg~\cite{luo2019pkuseg} toolkit to segment words. 
The vocabulary size is limited to $30,000$ for KaMed and MedDialog, and $20,000$ for MedDG. 
The lengths of the state text span and action text span are set to 10 and 3 in our experiments.
We employ the PCL-MedBERT\footnote{\url{https://code.ihub.org.cn/projects/1775/repository/mindspore\_pretrain\_bert}} embedding which is trained on a large-scale medical corpus.
We set the temperature of Gumbel-Softmax to $\tau= 3.0$, and anneal to $0.1$ in $30,000$ training steps.
We use the Adam optimizer \cite{kingma2014adam}; the learning rate is initialized to $1e^{-4}$ and decrease to $1e^{-5}$ gradually. 
For all models, we apply beam search decoding with a beam size of $5$ for response generation.

\begin{table}[t]
    \centering
    \small
    \caption{Automatic evaluation on the MedDG dataset. \textit{S-Sup} and \textit{A-Sup} indicate the supervision proportion of states and actions, respectively. Models that are able to use unlabeled data are marked with $^{\#}$.}
    \scalebox{0.93}{
    \begin{tabular}{l c c c c c c}
    \toprule
        \textbf{Model} & \textbf{S-sup} & \textbf{A-sup} & \textbf{ma-F1} & \textbf{mi-F1} & \textbf{EA} & \textbf{EG} \\
        \midrule
        \multirow{2}*{SEDST$^{\#}$} & 25\% & \multirow{2}*{0\%} & 13.50 & 20.94 & 22.39 & 33.19 \\
        ~ & 50\% & ~ & 12.29 & 19.97 & 23.12 & 33.59 \\ 
        \midrule
        \multirow{2}*{LABES$^{\#}$} & 25\% & \multirow{2}*{0\%} & 12.80 & 20.05 & 23.31 & 33.20 \\
        ~ & 50\% & ~ & 12.28 & 20.02 & 23.40 & 33.88 \\ 
        \midrule
        \multirow{3}*{NKD} & \multirow{3}*{0\%} & 25\% & 5.45 & 16.40 & 20.46 & 29.64 \\
        ~ & ~ & 50\% & 6.15 & 18.89 & 22.10 & 32.03 \\ 
        ~ & ~ & 100\% & 8.92 & 21.68 & 23.77 & 34.48 \\
        \midrule
        \multirow{3}*{PostKS$^{\#}$} & \multirow{3}*{0\%} & 25\% & 9.33 & 22.07 & 24.04 & 34.89 \\
        ~ & ~ & 50\% & 9.44 & 22.34 & 24.55 & 35.58 \\ 
        ~ & ~ & 100\% & 9.68 & 22.19 & 24.90 & 36.08 \\
        \midrule
        \multirow{4}*{HRED-Bert} & \multirow{4}*{0\%} & 10\% & 8.74 & 18.15 & 23.21 & 33.47 \\
        ~ & ~ & 25\% & 12.24 & 22.57 & 26.17 & 37.92 \\
        ~ & ~ & 50\% & 14.91 & 23.83 & 27.36 & 39.58 \\ 
        ~ & ~ & 100\% & 15.52 & 25.57 & 28.42 & 41.13 \\
        \midrule
        \multirow{3}*{DAMD} & 25\% & 25\% & 12.94 & 21.62 & 23.91 & 34.82 \\
        ~ & 50\% & 50\% & 14.26 & 23.70 & 24.83 & 36.06 \\ 
        ~ & 100\% & 100\% & 13.47 & 25.06 & 26.28 & 38.39 \\
        \midrule
        \multirow{3}*{MOSS$^{\#}$} & 25\% & 25\% & 12.74 & 23.03 & 25.83 & 37.46 \\
        ~ & 50\% & 50\% & 13.78 & 23.33 & 25.36 & 36.87 \\
        ~ & 100\% & 100\% & 13.84 & 24.36 & 25.21 & 36.69 \\
        \midrule
        \multirow{2}*{\OurMethod{}$\backslash$un}  & 25\% & 25\% & 10.86 & 20.74 & 24.49 & 35.69 \\
        ~ & 50\% & 50\% & 13.10 & 24.13 & 26.11 & 38.19 \\
        \midrule
        \multirow{6}*{\OurMethod{}$^{\#}$} & 25\% & 0\% & \textbf{15.79}\rlap{*} & \textbf{25.41}\rlap{*} & \textbf{24.09} & \textbf{35.29}\rlap{*} \\
        ~ & 50\% & 0\% & \textbf{14.75}\rlap{*} & \textbf{24.10}\rlap{*} & \textbf{25.69}\rlap{*} & \textbf{34.72}\rlap{*} \\
        ~ & 10\% & 10\% & \textbf{15.10}\rlap{*} & \textbf{24.85}\rlap{*} & \textbf{27.21}\rlap{*} & \textbf{39.72}\rlap{*} \\
        ~ & 25\% & 25\% & \textbf{15.88}\rlap{*} & \textbf{25.77}\rlap{*} & \textbf{27.73}\rlap{*} & \textbf{40.52}\rlap{*} \\  
        ~ & 50\% & 50\% & \textbf{14.72} & \textbf{26.63}\rlap{*} & \textbf{27.82}\rlap{*} & \textbf{40.69}\rlap{*} \\ 
        ~ & 100\% & 100\% & \textbf{15.31}\rlap{*} & \textbf{26.66}\rlap{*} & \textbf{27.51}\rlap{*} & \textbf{40.29}\rlap{*} \\
        \midrule
        \multirow{2}*{\OurMethod{}-Bert$^{\#}$} & 50\% & 50\% & \textbf{15.80} & \textbf{26.70}\rlap{*} & \textbf{28.21} & \textbf{41.18}\rlap{*} \\
        ~ & 100\% & 100\% & \textbf{16.11} & \textbf{27.60}\rlap{*} & \textbf{28.80} & \textbf{42.08} \\
        \bottomrule
    \end{tabular}
    }
    \label{tab:automatic-evaluation-supervised}
\end{table}


\smallnegskip
\section{Experimental Results}
\smallnegskip
\subsection{Overall performance}
\smallnegskip
We show the automatic evaluation results for all unsupervised models on KaMed and MedDialog in Table~\ref{tab:automatic-evaluation-unsupervised}, and the semi-supervised results in Table~\ref{tab:automatic-evaluation-supervised}. 
We see in Table~\ref{tab:automatic-evaluation-unsupervised} that \OurMethod{} significantly outperforms all baselines in terms of most evaluation metrics on both datasets. 
In terms of D@1 and D@2 \OurMethod{} outperforms other baselines as the generated responses in \OurMethod{} are more diverse. 
For KaMed, \OurMethod{} achieves an increase of 14.68\%, 36.81\%, 61.00\%, and 67.57\% over PostKS in terms of B@2, R@2, D@1, and D@2, respectively.
For MedDialog, \OurMethod{} gives an increase of 21.47\%, 14.17\%, 31.29\%, and 43.63\% over PostKS.
Models without reasoning give high ma-P and mi-P scores, but they do not perform well in terms of ma-R, mi-R, ma-F1, mi-F1.
In terms of ma-R, mi-R, ma-F1, mi-F1, EA and EE, \OurMethod{} outperforms all baselines by a large margin. 
Hence, \OurMethod{} is effective in predicting physician actions.
Table~\ref{tab:automatic-evaluation-supervised} shows the performance in semi-supervised settings on the MedDG dataset.
SEDST and LABES are two state-of-the-art semi-supervised state tracking approaches. 
Without labeled action data, \OurMethod{} still achieves 23.35\% and 26.73\% improvements over LABES in terms of ma-F1 and mi-F1 with 25\% labeled states; when the state labeling proportion increases to 50\%, \OurMethod{} achieves an increase of 20.11\% and 20.38\%. 
\OurMethod{} outperforms \OurMethod{}\textbackslash un by 12.36\% and 10.36\% in terms of ma-F1 and mi-F1 with the supervision proportion set to 50\%; the increase is more significant with a lower supervision proportion. Thus, unlabeled data improves the performance of \OurMethod{}.
\OurMethod{} outperforms MOSS by a large margin despite the fact that MOSS can also use unlabeled data; it outperforms MOSS by 11.90\% and 14.14\% in terms of mi-F1 when with 25\% and 50\% labeled data, respectively.
\OurMethod{} significantly outperforms HRED-Bert in terms of all metrics when the supervision proportion $\le$ 25\%. 
With 50\% and 100\% annotated data, \OurMethod{}-Bert still outperforms HRED-Bert by 12.04\% and 7.93\%  in terms of mi-F1. 

In Table~\ref{tab:human-evaluation}, we perform a human evaluation on the KaMed and MedDG dataset to investigate the unsupervised and semi-supervised performance of \OurMethod{}. 
\OurMethod{} achieves the best performance in terms of all metrics on both datasets. 
On KaMed, \OurMethod{} outperforms SEDST and PostKS in terms of KC and EQ by a large margin. The result is consistent with our automatic evaluation results, confirming the importance of simultaneously modeling patient state and physician action.
On MedDG, MOSS slightly outperforms DAMD, a fully-supervised method. \OurMethod{} achieves a 13\% and 15\% increase over MOSS in terms of KC and EQ. 
Thus, the unlabeled states and actions inferred by \OurMethod{} help to improve performance.
We compute the average pairwise Cohen's kappa ($\kappa$) to measure the consistency between annotators, and find that $0.6 \geq \kappa \geq 0.4$ for all metrics.

\negskip
\subsection{Ablation study}
As shown in Table~\ref{tab:\OurMethod{}-ablation-study}, all components in \OurMethod{} contribute to its performance.
On KaMed, the performance of \OurMethod{}$\backslash$S and \OurMethod{}$\backslash$A drop by 42.27\% and 17.64\% in terms of mi-F1 respectively. 
On MedDialog, \OurMethod{}$\backslash$S and \OurMethod{}$\backslash$A drop by 50.54\% and 44.76\% respectively, which means that states and actions are equally important, modeling only one of them is far from enough.
The performance of \OurMethod{}$\backslash$C drops sharply in terms of all metrics; it drops by 22.32\% and 58.85\% in terms of mi-F1 on KaMed and MedDialog, respectively. 
\OurMethod{}$\backslash$G drops a little, 3.66\% and 1.34\% in terms of mi-F1 on KaMed and MedDialog. 

The context-reasoning detector is able to leverage the raw dialogue to improve the reasoning ability, whereas the graph-reasoning detector can only use prior knowledge in the knowledge base. 
In terms of EA and EG, \OurMethod{}$\backslash$C outperforms \OurMethod{}$\backslash$A by 18.12\% and 16.98\% on KaMed, 28.03\% and 26.27\% on MedDialog, despite the fact that the mi-F1 score is close to \OurMethod{}$\backslash$A. 
Hence, \OurMethod{}$\backslash$C benefits from the rich semantics of external knowledge though the entity name in the knowledge graph does not strictly match the dialogue corpus. 
Without the 2-stage collapsed inference training trick (that is, \OurMethod{}$\backslash$2s), the mi-F1 score decreases by 18.22\% and 18.37\% on KaMed and MedDialog, respectively.

\begin{table}
    \centering
    \small
    \caption{Human evaluation on the KaMed and MedDG datasets (with 25\% annotations).}
    \label{tab:human-evaluation}
    \scalebox{1.0}{
    \begin{tabular}{l r r r c@{\hspace*{5mm}} l r r r}
    \toprule
        \multirow{2}*{\textbf{Model}} & \multicolumn{3}{c}{\textbf{\# KaMed}} && \multirow{2}*{\textbf{Model}} & \multicolumn{3}{c}{\textbf{\# MedDG}} \\
        \cmidrule(r){2-4} \cmidrule{7-9}
        ~ & \textbf{Flu} & \textbf{KC} & \textbf{EQ} && ~ & \textbf{Flu} & \textbf{KC} & \textbf{EQ}\\
        \midrule
        SEDST & 3.52 & 1.88 & 1.81  && DAMD & 3.77 & 2.62 & 2.49 \\ 
        PostKS & 3.20 & 1.77 & 1.67  && MOSS & 3.76 & 2.88 & 2.59 \\
        \OurMethod{} & \textbf{4.21} & \textbf{2.96} & \textbf{2.69} && \OurMethod{} & \textbf{4.00} & \textbf{3.26} & \textbf{2.99} \\ 
        \midrule
        $\kappa$ & 0.54 & 0.56 & 0.49 && $\kappa$ & 0.45 & 0.47 & 0.48 \\
    \bottomrule
    \end{tabular}
    }
\end{table}

\begin{table}[t]
    \centering
    \small
    \caption{Ablation study: A comparison of different variations by masking out specific sub-module.}
    \scalebox{1.0}{
    \begin{tabular}{l c c c c c c}
    \toprule
        \multirow{2}*{\textbf{Model}} & \multicolumn{3}{c}{\textbf{\# KaMed}} & \multicolumn{3}{c}{\textbf{\# MedDialog}} \\
        \cmidrule(r){2-4} \cmidrule(r){5-7}
        ~ & \textbf{mi-F1} & \textbf{EA} & \textbf{EG} & \textbf{mi-F1} & \textbf{EA} & \textbf{EG} \\ \midrule
        \OurMethod{}$\backslash$S & 15.14 & 30.01 & 29.64 & 12.03 & 21.48 & 22.05 \\
        \OurMethod{}$\backslash$A & 18.31 & 31.67 & 31.67 & 12.51 & 21.26 & 22.53 \\
        \OurMethod{}$\backslash$G & 20.78 & 42.02 & 41.84 & 17.87 & 33.09 & 35.41 \\
        \OurMethod{}$\backslash$C & 17.61 & 37.41 & 37.05 & 11.40 & 27.22 & 28.45 \\
        \OurMethod{}$\backslash$2s  & 18.22 & 38.09 & 37.91 & 15.30 & 28.93 & 30.82 \\
        \midrule
        \OurMethod{} & \textbf{21.54} & \textbf{42.37} & \textbf{42.33} & \textbf{18.11} & \textbf{34.51} & \textbf{36.79} \\
        \bottomrule
    \end{tabular}
    }
    \label{tab:\OurMethod{}-ablation-study}
\end{table}

\negskip
\subsection{Impact of $|S|$ and $|A|$}
\begin{figure}[t]
    \centering
    \includegraphics[width=1.0\linewidth]{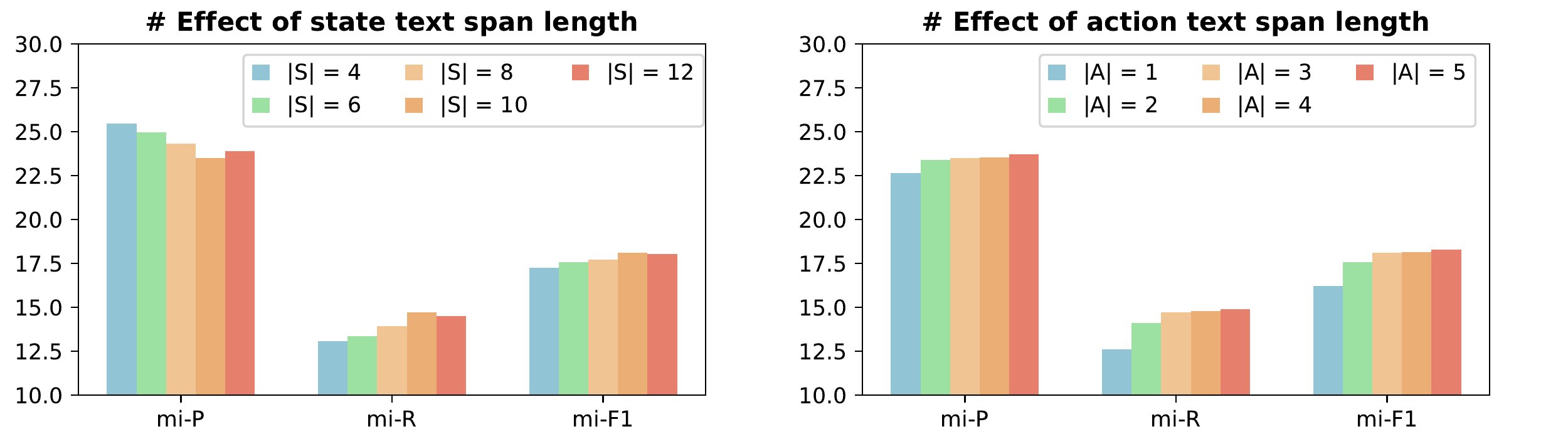}
    \caption{The effect of text span length on MedDialog}
    \label{fig:comparision-with-different-a-s-len}
\end{figure}

The length of state and action text span are set to fixed integers $|S|$ and $|A|$ respectively, as they could not be inferred in unsupervised learning. 
We conduct experiments on MedDialog by setting $|S|$ to values in $\{4, 6, 8, 10, 12\}$ while fixing $|A|$ to 3, and selecting $|A|$ from $\{1, 2, 3, 4, 5\}$ while fixing $|S|$ to 10, to see the effects of $|S|$ and $|A|$.
The results are shown in Fig.~\ref{fig:comparision-with-different-a-s-len}.
Focusing on the left part, we see that mi-P decreases, while mi-R and mi-F1 increase as the state text span length grows.
As $|S|$ increases from 4 to 10, the mi-R and mi-F1 achieve 12.79\% and 4.92\% improvements, while mi-P decreases by 6.53\%. 
On the right side, we see a tendency for all metrics to increase as $|A|$ increases, and the upward trend gradually slows down. 
As $|A|$ increases from 1 to 3, \OurMethod{} achieves 3.71\%, 16.81\% and 11.72\% improvements in terms of mi-P, mi-R and mi-F1. 
The recall score rises a lot as a longer action text spans are able to present more information in the reply. 
As $|A|$ further increases from 3 to 5, the improvements are relatively small, i.e., 1.04\% in terms of mi-F1. We have qualitatively similar findings on the KaMed dataset, which we omit due to space limitations.

\negskip
\subsection{Explainability comparsion}
To explicitly assess the explainability of \OurMethod{}'s results, we calculate MEP and MER scores of action text spans in KaMed. 
We take a random sample of 50 dialogues from KaMed and manually compare the explainability performance of \OurMethod{} and PostKS. 
The results are listed in Table~\ref{tab:explainability-comparision}. 
We observe that \OurMethod{} outperforms PostKS by a large margin in terms of MEP and MER; our user study also shows that \OurMethod{} achieves a 44\% win rate. 
This confirms that \OurMethod{} can provide more interpretable results in responses and action text spans.

\negskip
\subsection{Case study}
We randomly sample an example from the KaMed test set to compare the performance of \OurMethod{}, SEDST and PostKS in Tab.~\ref{tab:case_in_english}. 
The dialogue occurs in the ear-nose-throat department and concerns the treatment of `\textit{allergic rhinitis}'. 
In the $3$rd round we see that SEDST and \OurMethod{} can both generate a state text span (i.e., $S_3$ in Tab.~\ref{tab:case_in_english}) to model the patient state. 
\OurMethod{} tracks patient disease and symptoms `\textit{allergic rhinitis, stuffy nose, sneezing}', then prescribes the correct drugs `\textit{Nasonex}' and `\textit{Montelukast}' to meet the patient requirements (it is correct though does not match the gold response); 
We see a reasoning path `allergic rhinitis \ding{222} treated\_by \ding{222} Montelukast ($0.09$)' in the graph, where $0.09$ indicates the copying weight of `Montelukast' in the graph reasoning detector.
SEDST and PostKS both fail to generate an accurate and interpretable response; this confirms the importance of simultaneously modeling patient states and physician actions.
\OurMethod{} is able to generate interpretable responses with explicit text spans and reasoning paths.
\negskip

\begin{table}[t]
    \centering
    \small
    \caption{Explainability comparison of \OurMethod{} and PostKS.}
    \scalebox{1.0}{
    \begin{tabular}{l c c c}
    \toprule
        \textbf{Model} & \textbf{MEP} & \textbf{MER} & \textbf{Win Rate} (User Study) \\
        \midrule
        PostKS  & 30.19 & 56.48 & 16.00 \\
        \OurMethod{} & \textbf{44.22} & \textbf{82.61} & \textbf{44.00} \\
        \bottomrule
    \end{tabular}
    }
    \label{tab:explainability-comparision}
\end{table}

\begin{table}[!htbp]
\vspace*{-1mm}
\scriptsize
    \centering
    \caption{One case extracted from KaMed, \xmark and \cmark denote that the response is incorrect and excellent, respectively. }
    \label{tab:case_in_english}
    \begin{tabularx}{\linewidth}{lX}
    \toprule
    $U_1$ : & Is it allergic rhinitis (female, 19 years old)? My nose is very itchy and runny after running. \\
    $R_1$: & How long has it been? Did you sneeze? Stuffy nose? \\
    $U_{2}$ : & Um. I usually feel the nose be a little uncomfortable, often dry nose. \\
    $R_{2}$: & You can use Budesonide Nasal Spray. \\
    $U_3$ : & Do you have any other suggestions for my symptoms?\\
    \midrule
    Golden : & Flushed your nasal cavity with physiological seawater, take loratadine tablets. \\
    \midrule
    $S_{3}$: & allergic rhinitis, stuffy nose, sneezing, Budesonide Nasal Spray \\
    SEDST: & Spray your nose with Budesonide Nasal Spray. \xmark \\
    \midrule
    $A_3$: & allergic rhinitis \\
    PostKS: & Your symptom is caused by allergic rhinitis, suggest you go to the hospital to check the nose. \xmark \\
    \midrule
    $S_{3}$ : & allergic rhinitis, stuffy nose, sneezing \\
    $G^{local}_n$: & allergic rhinitis \ding{222} treated\_by \ding{222} Montelukast (0.09) ; allergic rhinitis \ding{222} treated\_by \ding{222} Cetirizine (0.04); allergic rhinitis \ding{222} treated\_by \ding{222} Dexamethasone (0.02) \\
    $A_{3}$ : & prescribe medicine, Nasonex, Montelukast \\
    \OurMethod{}: &Spray your nose with Nasonex , take Montelukast Sodium Chewable Tablets. \cmark \\
    \bottomrule
    \end{tabularx}
    \vspace*{-3.92mm}
\end{table}


\vspace*{-3mm}
\section{conclusions}
\label{sec:conclusion}

In this paper, we focus on medical dialogue response generation with a large-scale unlabeled corpus. 
We propose a generative model named \OurMethod{}, which uses latent variables to model unobserved patient state and physician actions. 
We derive the ELBO for \OurMethod{} and propose a 2-stage collapsed inference training trick that decomposes the ELBO into two learning objectives.
Extensive experiments on three medical dialogue datasets show that \OurMethod{} achieves state-of-the-art performance on both unsupervised and semi-supervised learning. 
Furthermore, in a fully-supervised setting, \OurMethod{}-Bert which is a variation of \OurMethod{} augmented by Bert achieves the best results compared to all the baselines.
Analysis also confirms that \OurMethod{} is able to generate interpretable results.

\OurMethod{} proves the value of having a large-scale unlabeled medical corpus. 
It can be also applied to other task-oriented dialogue systems with few annotated data.
As to our future work, we aim to leverage the labeled data of a single hospital department to improve the \ac{MDG} performance on other departments without labeled data by transfer learning or zero-shot learning. 

\vspace*{-1.5mm}
\section*{Reproducibility}
\smallnegskip
Our code and dataset are available at \url{https://github.com/lddsdu/VRBot}.

\vspace*{-2mm}
\begin{acks}
\smallnegskip
This work was supported by the Natural Science Foundation of China (61902219, 61972234, 62072279), the National Key R\&D Program of China with grant No.\ 2020YFB1406704, 
the Key Scientific and Technological Innovation Program of Shandong Province (2019JZZY010129), 
Shandong University multidisciplinary research and innovation team of young scholars (No.\ 2020QNQT017),
the Tencent WeChat Rhino-Bird Focused Research Program (JR-WXG-2021411), the Hybrid Intelligence Center, 
and
a 10-year program funded by the Dutch Ministry of Education, Culture and Science through 
the Netherlands Organisation for Scientific Research, \url{https://hybrid-intelligence-centre.nl}. 
All content represents the opinion of the authors, which is not necessarily shared or endorsed by their respective employers and/or sponsors.
\end{acks}

\newpage
\bibliographystyle{ACM-Reference-Format}
\balance
\bibliography{references}

\end{sloppypar}
\end{document}